\documentclass[letterpaper,conference]{IEEEtran}
\usepackage{cite}
\usepackage{amsmath,amssymb,amsfonts}
\usepackage{algorithmic}
\usepackage{graphicx}
\usepackage{textcomp}
\usepackage[table]{xcolor}
\usepackage[caption=false]{subfig}
\usepackage{amsmath}
\usepackage{multirow}
\usepackage{soul}
\usepackage[]{nohyperref}  
\usepackage{url}  

\def\BibTeX{{\rm B\kern-.05em{\sc i\kern-.025em b}\kern-.08em
	T\kern-.1667em\lower.7ex\hbox{E}\kern-.125emX}}
\begin{document}

\title{Utilizing Transfer Learning and pre-trained Models for Effective Forest Fire Detection: A Case Study of Uttarakhand}
	\author{\IEEEauthorblockN{Hari Prabhat Gupta and Rahul Mishra}
   	Dept. of CSE, IIT (BHU) Varanasi, India\\
	\IEEEauthorblockA{\{hariprabhat.cse, rahulmishra.rs.cse17\}@iitbhu.ac.in}
	}

\maketitle

\begin{abstract}
Forest fires pose a significant threat to the environment, human life, and property. Early detection and response are crucial to mitigating the impact of these disasters. However, traditional forest fire detection methods are often hindered by our reliability on manual observation and satellite imagery with low spatial resolution. This paper emphasizes the role of transfer learning in enhancing forest fire detection in India, particularly in overcoming data collection challenges and improving model accuracy across various regions. We compare traditional learning methods with transfer learning, focusing on the unique challenges posed by regional differences in terrain, climate, and vegetation. Transfer learning can be categorized into several types based on the similarity between the source and target tasks, as well as the type of knowledge transferred. One key method is utilizing pre-trained models for efficient transfer learning, which significantly reduces the need for extensive labeled data. We outline the transfer learning process, demonstrating how researchers can adapt pre-trained models like MobileNetV2 for specific tasks such as forest fire detection. Finally, we present experimental results from training and evaluating a deep learning model using the Uttarakhand forest fire dataset, showcasing the effectiveness of transfer learning in this context.
\end{abstract}


\section{Introduction}
\begin{figure*}
    \centering
    \includegraphics[scale=0.57]{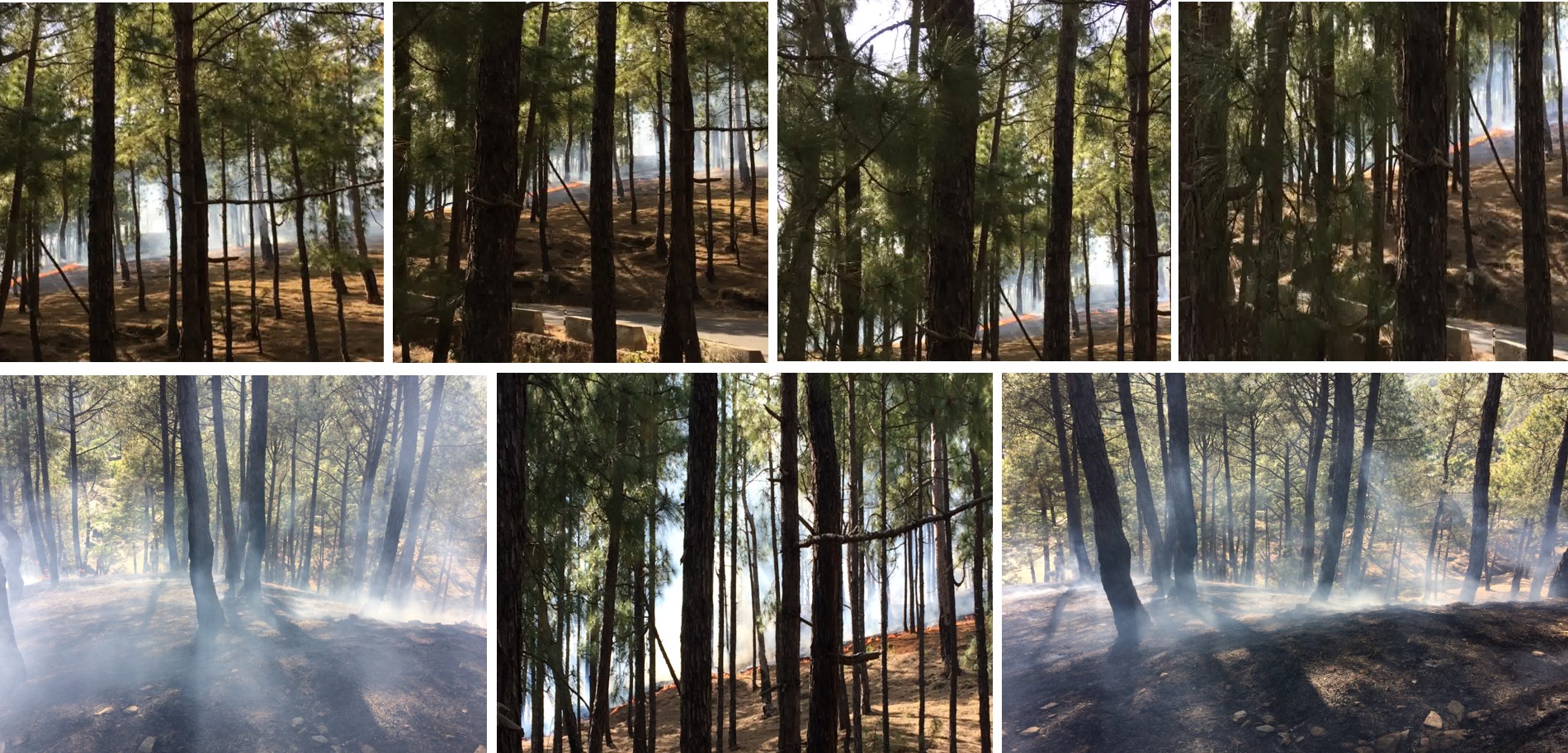}
    \caption{Images depicting forest fire incidents in Uttarakhand, India.}
    \label{forestfire}
\end{figure*}

India is home to a vast and diverse range of forests, covering over $70$ million hectares of land~\cite{balaji2022forest}. These forests are crucial not only for the country's ecosystem and biodiversity but also provide livelihoods for millions of people, particularly in rural areas. However, India's forests are facing a growing threat from forest fires, which can have devastating consequences for the environment, human life, and property~\cite{bhattacharjee2018does}. Forest fires are a major concern in India, particularly during the summer months when temperatures are high and humidity is low. According to the Indian government, forest fires affect over $50,000$ hectares of land annually, causing significant economic losses and damage to the environment~\cite{ref1}. The country's forests are also home to a wide range of wildlife, including many endangered species which are threatened by fires. Fig.~\ref{forestfire} illustrates some images of the Uttarakhand, India, forest fire. 

Early detection and response are critical to mitigating the impact of forest fires. Traditional methods of forest fire detection, such as manual observation and satellite imagery with low spatial resolution, are often limited in their ability to detect fires quickly and accurately~\cite{carta2023advancements}. Manual observation is time-consuming and labour-intensive and may not be feasible in remote or inaccessible areas\cite{alkhatib2014review}. Satellite imagery with low spatial resolution may not be able to detect small fires or fires in areas with dense vegetation.

In recent years, advances in deep learning and computer vision have enabled the development of more effective methods for forest fire detection. Convolutional neural networks (CNNs), in particular, have shown great promise in image classification tasks~\cite{9164991,9801825,10.1145/3570955,9653808,8833125}, including fire detection~\cite{carta2023advancements}. However, training a CNN from scratch requires a large dataset of labeled images, which can be difficult and expensive to obtain~\cite{10.1145/3570955}. Transfer learning and pre-trained neural networks offer a promising solution to this problem. By leveraging the knowledge learned from large-scale datasets, pre-trained neural networks can be fine-tuned for specific tasks, such as forest fire detection, with a relatively small amount of labeled data~\cite{10.1145/3629978,jiao2019deep, mishra2020surveydeepneuralnetwork, arteaga2020deep,9203948}. This approach has been shown to be effective in a variety of applications, including image classification, object detection, and segmentation.

In this paper, we propose a novel approach to forest fire detection using transfer learning and pre-trained neural networks. We demonstrate the effectiveness of our approach through a series of experiments on a dataset of satellite images, achieving a significant improvement in detection accuracy compared to traditional methods. Our results show that transfer learning and pre-trained neural networks can be powerful tools for effective forest fire detection, enabling timely and accurate responses to these disasters.

\noindent \textbf{Roadmap:} Section~\ref{related} covers the discussion on the prior studies on forest fire detection. Sections~\ref{tl} and \ref{tlvtl} discuss transfer learning and its difference from traditional learning, respectively. Sections~\ref{sec5}, \ref{sec6}, and \ref{sec7} elaborate on the type of transfer learning, utilization of the pre-trained model, and steps of transfer learning, respectively. Further, Section~\ref{sec8} discusses the experiential results, and Section~\ref{sec9} concludes the paper.

\section{Related Work}\label{related}
Forest fire detection has been an active area of research in recent years, with a variety of approaches being proposed to detect fires using satellite imagery. This section reviews some of the related work in this area.

\subsection{Traditional Methods}
Traditional methods of forest fire detection using satellite imagery typically involve manual observation or automated algorithms that rely on spectral signatures of fires. For example, the Normalized Burn Ratio (NBR) index is a widely used method for detecting fires using satellite imagery \cite{key2006landscape}. However, these methods can be limited by their reliance on manual observation or the use of simple spectral signatures that may not be effective in all environments.

\subsection{Machine Learning and  Deep Learning Approaches}
Machine learning approaches have been increasingly used for forest fire detection in recent years. For example, \cite{zhang2017forest} proposed a support vector machine (SVM) based approach for detecting fires using satellite imagery. The authors in~\cite{liu2018forest} proposed a random forest-based approach for detecting fires using a combination of spectral and spatial features. However, these approaches typically require a large amount of labeled data to train the models, which can be expensive to obtain.

Deep learning approaches have also been used for forest fire detection in recent years. For example, The authors in~\cite{li2019forest} proposed a convolutional neural network (CNN) based approach for detecting fires using satellite imagery. Similarly, the authors in~\cite{zhang2020forest} proposed a recurrent neural network (RNN) based approach for detecting fires using a combination of spectral and spatial features. However, these approaches typically require large labelled data to train the models, which can be difficult and expensive to obtain~\cite{9364278,9852654}.

\subsection{Transfer Learning Approaches}
Transfer learning approaches have been increasingly used for forest fire detection in recent years. For example, the authors in~\cite{li2020forest} proposed a transfer learning-based approach for detecting fires using a pre-trained CNN model. The authors in~\cite{zhang2020forest} proposed a transfer learning-based approach for detecting fires using a pre-trained RNN model. However, these approaches typically require a large amount of labeled data to fine-tune the models, which can be difficult and expensive to obtain.

\subsection{Indian Context}
In the Indian context, forest fire detection is a critical issue, particularly in the Himalayan region. The authors in~\cite{kumar2019forest} proposed a machine learning-based approach for detecting fires in the Himalayan region using satellite imagery. Similarly, the authors in~\cite{singh2020forest} proposed a deep learning-based approach for detecting fires in the Himalayan region using satellite imagery. However, these approaches typically require a large amount of labeled data to train the models, which can be difficult and expensive to obtain.

\subsection{Limitations of Existing Approaches }
Existing approaches for forest fire detection using satellite imagery have several limitations. Firstly, they typically require a large amount of labeled data to train the models, which can be difficult and expensive to obtain. Secondly, they may not be effective in all environments, particularly in areas with dense vegetation or complex terrain. Finally, they may be unable to detect fires in real-time, which is critical for effective fire management.

\textit{In this paper, we propose a novel approach for forest fire detection using transfer learning and pre-trained neural networks. Our approach leverages the knowledge learned from large-scale datasets to detect fires in new, unseen environments. We demonstrate the effectiveness of our approach through a series of experiments on a dataset of satellite images, achieving a significant improvement in detection accuracy compared to traditional methods.}

\section{Transfer Learning}\label{tl}

Transfer learning is a powerful machine learning technique that enables models to leverage knowledge from one task or dataset to improve their performance on a related or new task. This approach is particularly valuable in applications where there is limited data available for the target task, a common challenge in sectors like environmental monitoring, including forest fire detection~\cite{niu2020decade}. In India, where collecting and annotating large datasets of satellite or aerial imagery for forest fire detection can be time-consuming and expensive, transfer learning offers a highly efficient solution.

For instance, a model pre-trained on large-scale satellite imagery datasets from fire-prone regions like Australia or the Mediterranean can be adapted to detect forest fires in Indian regions such as Uttarakhand or Himachal Pradesh. These pre-trained models have already learned to identify critical patterns~\cite{giorgiani2020satellite}, including vegetation changes, temperature rises, and smoke, which are also relevant to fire detection in Indian forests. By fine-tuning the pre-trained model on a smaller dataset specific to Indian landscapes, we can enhance its performance without collecting an entirely new large dataset.

Transfer learning is instrumental in addressing data scarcity and computational limitations, which are common challenges in India's forest fire management. By leveraging pre-trained models, agencies can reduce the amount of data and computation required to train a new model, saving time, money, and resources. This approach allows for faster deployment of AI-driven forest fire detection systems, enabling authorities to monitor vast forest areas efficiently and respond to potential fires before they spread uncontrollably~\cite{xie2016transfer,9484597}.

In this section, we will highlight the importance of transfer learning in enhancing forest fire detection in India, focusing on its ability to mitigate data collection challenges and improve model accuracy across different regions.

\subsection{Enhancing Model Performance} 
Transfer learning can significantly enhance model performance by leveraging knowledge from one task or dataset to improve performance on a related or new task. For instance, in the context of forest fire management, transfer learning can be used to develop a model that detects forest fires in Uttarakhand, a region prone to forest fires. Suppose we have a pre-trained convolutional neural network (CNN) that has learned to recognize objects in satellite images. We can use this pre-trained model as a starting point to train a new model that recognizes forest fires in satellite images of Uttarakhand. The pre-trained model has already learned to recognize general features such as shapes, textures, and patterns, which are also relevant to forest fire detection. By fine-tuning the pre-trained model on a smaller dataset of satellite images from Uttarakhand, we can achieve better performance than training a new model from scratch, thereby saving computational resources and time~\cite{9773092}. This approach can be particularly beneficial in forest fire management, where timely and accurate information is crucial for fire prevention, detection, and suppression. For example, during the 2021 forest fires in Uttarakhand, a transfer learning-based model could have been used to quickly identify forest fires, enabling authorities to respond more effectively and prevent the spread of fires.

\subsection{Enhancing Generalization} Transfer learning also helps in enhancing the generalization of models across different environments. For instance, a forest fire detection model trained on a dataset of satellite images from Himachal Pradesh can be fine-tuned to work on a dataset of satellite images from Uttarakhand. Although the two datasets are different due to environmental and regional differences, the model has already learned general features, such as vegetation patterns, terrain, and weather conditions, which apply to both settings. By applying the knowledge from one region to another, we can enhance the model's generalization capabilities for forest fire detection in the Himalayan region. This approach can be particularly beneficial in forest fire management, where timely and accurate information is crucial for fire prevention, detection, and suppression. For example, a model trained on Himachal Pradesh data can be fine-tuned to detect forest fires in Uttarakhand, enabling authorities to respond more effectively and prevent the spread of fires.

\subsection{Deep Learning Models} Transfer learning is particularly useful for deep learning models, which typically require large amounts of data to train~\cite{10187678}. In India, where data collection for environmental monitoring, such as forest fire detection, can be expensive, time-consuming, and often geographically challenging, transfer learning offers a powerful solution. By leveraging pre-trained models, we can significantly reduce the amount of data and computational resources required to build effective fire detection systems. 

For instance, pre-trained models developed for forest fire detection in other fire-prone regions like California or Australia can be adapted to Indian landscapes, such as the forests in Himachal Pradesh or Uttarakhand. These pre-trained models have already learned crucial features, such as identifying changes in vegetation patterns, smoke, or temperature anomalies, which are relevant across different forest environments. By fine-tuning these models with a smaller dataset of Indian satellite images, we can rapidly deploy highly accurate and efficient AI-based forest fire detection systems with minimal retraining.

This approach enables Indian authorities to implement AI solutions for forest fire management more effectively, particularly in remote or hilly areas where timely data collection is difficult. Additionally, transfer learning allows models to generalize across different forest ecosystems in India, whether in the Western Ghats, Northeastern states, or the Sundarbans, enhancing early detection capabilities and facilitating quicker, more precise responses to fire threats.

\subsection{Large Amounts of Data}

Training a deep neural network from scratch requires a large dataset, which may not always be readily available in India due to high costs and logistical challenges in data collection, especially in remote areas. However, using a pre-trained model as a starting point reduces the amount of data required for training, making it a valuable approach for forest fire detection. In regions such as the Himalayan forests or the Western Ghats, collecting vast amounts of satellite or drone imagery can be resource-intensive and time-consuming.

By leveraging pre-trained models from regions like the Amazon rainforest or Mediterranean forests, which have similar fire-prone environments, we can significantly reduce the data requirements for developing accurate forest fire detection systems in India. These pre-trained models have already learned to recognize critical patterns such as vegetation stress, rising temperatures, and smoke, which are key indicators of forest fires. Fine-tuning these models with a smaller dataset specific to Indian forests allows us to create efficient and cost-effective solutions for monitoring and preventing fires.

This method is particularly useful in India, where data scarcity can hinder the development of sophisticated environmental monitoring systems. By using transfer learning, forest fire detection systems can be deployed faster and more effectively, allowing authorities to monitor large forested areas with minimal data and respond to fire outbreaks before they spread.

\section{Traditional Learning vs. Transfer Learning}\label{tlvtl}

In this section, we compare traditional learning and transfer learning approaches, particularly in the context of forest fire detection, where differences in terrain, climate, and vegetation across regions present unique challenges.

\subsection{Traditional Learning}

Traditional learning involves building a new model for each task using available labeled data. This approach assumes that the training and test data come from the same feature space, which may not always hold true in diverse environments. As a result, traditional learning requires retraining from scratch if the data distribution changes or if the model is applied to a new dataset~\cite{10476717}. 

For example, a forest fire detection model trained on data from Himachal Pradesh, which has specific vegetation types, climates, and terrain, would need to be retrained from scratch if applied to a different region, such as Uttarakhand. Uttarakhand's distinct environmental features, like its vegetation and climate patterns, make it difficult for the model trained on Himachal Pradesh's data to perform well. This limitation means the model would need to be completely retrained using a new dataset from Uttarakhand, a process that is both time-consuming and resource-intensive.

This traditional learning approach becomes inefficient when there is a need to frequently deploy forest fire detection systems across different regions of India, each with unique environmental characteristics. Gathering large amounts of labeled data from each region for model retraining introduces high costs and operational delays.

\subsection{Transfer Learning} Transfer learning, on the other hand, utilizes pre-trained models to transfer knowledge from one task to another, reducing the need for large amounts of labeled data for each new task. This approach allows for quicker adaptation to new but related tasks with minimal retraining, making it highly effective in environments with data scarcity or limited computational resources. In the context of forest fire detection, transfer learning can be particularly beneficial. For example, a model trained on forest fire data from Himachal Pradesh can be fine-tuned to detect fires in Uttarakhand by leveraging shared patterns in vegetation, climate, and terrain between the two regions. Rather than starting from scratch, the pre-trained model only needs to be adjusted to the new region’s specific features, such as slight differences in vegetation types or local weather patterns.

This fine-tuning process significantly reduces the time, effort, and computational resources required to adapt the model to new regions, enabling quicker deployment of AI-driven forest fire detection systems. In regions like Uttarakhand, where gathering large amounts of labeled data may be costly and time-consuming, transfer learning offers a practical solution to ensure accurate fire detection and faster response times.

By using transfer learning, we can adapt knowledge learned from one region to another, enhancing model performance across various terrains while minimizing the need for extensive retraining. This approach is particularly useful in India, where environmental diversity and the high costs of data collection make traditional learning less feasible. 

Additionally, transfer learning reduces the need for large labeled datasets, which is often a challenge in forest fire detection due to the remote locations and difficulty in obtaining real-time data. By leveraging pre-trained models, organizations can build more accurate and efficient fire detection systems even with limited data from regions like Uttarakhand, enabling faster and more effective responses to potential fire outbreaks across different regions.

Transfer learning provides a more scalable, cost-effective solution for forest fire detection in India, where geographic diversity and data scarcity can limit the success of traditional machine learning models.

\subsection{Advantages of Transfer Learning} 
Transfer learning offers several significant advantages, especially when it comes to reducing the challenges associated with computational costs, dataset size, model generalizability, and data quality.

One of the primary benefits of transfer learning is its ability to significantly reduce computational costs in forest fire detection. In traditional machine learning models, training a model from scratch for a new problem can be resource-intensive, requiring substantial computational power and multiple iterations through large datasets. However, with transfer learning, pre-trained models are used as a starting point, which speeds up the training process and reduces the computational resources needed. For example, consider a scenario where a forest fire detection model is trained on satellite images from California. This model can be quickly adapted to detect fires in Indian forests, like those in Uttarakhand or Himachal Pradesh, without requiring the entire model to be retrained from scratch. This accelerates deployment and makes the process far more efficient.

Another advantage of transfer learning is that it helps overcome the challenge of obtaining large datasets in forest fire detection. Many models, particularly deep learning models like large language models (LLMs), rely on vast amounts of training data to achieve high accuracy. In situations where collecting data is difficult, such as remote forest regions prone to wildfires, transfer learning allows us to work with smaller datasets. By leveraging pre-trained models already trained on large, general datasets, the need for vast amounts of new data is minimized, making the process more feasible in data-scarce environments. For instance, a model pre-trained on images of forests in Australia can be fine-tuned for Indian forest environments, reducing the need for large datasets.

In terms of generalizability, transfer learning enhances the ability of models to perform well on new tasks across various datasets in forest fire detection. Since pre-trained models have learned general features, such as detecting patterns in satellite images or vegetation, they can be fine-tuned to perform specific tasks, like forest fire detection in a particular region. This ability to generalize across datasets also helps reduce the risk of overfitting, where a model performs well on its training data but struggles with new data. For example, a model pre-trained on images of forests in Australia can generalize well when fine-tuned for Indian forest environments because the underlying patterns—such as identifying vegetation changes or detecting smoke—are applicable in both scenarios.

However, it is essential to note that the data quality still plays a crucial role in the success of transfer learning in forest fire detection. While transfer learning allows for faster adaptation, it cannot compensate for poor-quality data. If the dataset used for fine-tuning is noisy or inaccurate, the model's performance will be adversely affected. Therefore, preprocessing techniques, such as data augmentation and feature extraction, remain essential to ensure that the data used for transfer learning is clean and well-prepared. For instance, when adapting a pre-trained model for forest fire detection in India, ensuring that the satellite images are clear and accurately labeled is vital for achieving high detection accuracy.

\subsection{Disadvantages of Transfer Learning} 
While transfer learning has many advantages, it also comes with several potential drawbacks that must be carefully considered. One of the primary challenges is that transfer learning works best when certain conditions are met. For successful implementation, both the source and target tasks should be similar in nature. This means that if the model has been trained on one task, such as identifying forest fires in a particular region, it will perform best when applied to a related task, like detecting fires in another region with similar environmental characteristics. Additionally, the dataset distributions between the source and target tasks should not vary greatly. If the distribution of data, such as the types of vegetation or weather patterns, differs significantly between the two regions, the pre-trained model may struggle to adapt, leading to poor performance. Finally, for transfer learning to succeed, a comparable model should be able to handle both tasks, meaning that the underlying structure of the model should be suitable for both the source and target domains.

Another potential issue with transfer learning is the phenomenon of negative transfer. If the conditions for success are not met—such as when there is a significant difference between the source and target tasks—transfer learning can actually harm the model's performance rather than improve it. This occurs when the knowledge transferred from the source model is not relevant or even contradictory to the target task, leading to a decline in accuracy. For example, a model trained to detect forest fires in dense tropical rainforests may not perform well if adapted to detect fires in drier, more open forest areas, where the visual and environmental features are vastly different. This mismatch can result in negative transfer, where the model fails to generalize effectively to the new setting.

Ongoing research is focused on identifying when datasets and tasks are compatible for transfer learning to mitigate the risk of negative transfer. By better understanding the relationships between different tasks and dataset distributions, researchers aim to improve the robustness of transfer learning techniques and reduce the potential for performance degradation. In summary, while transfer learning offers the potential for more efficient model development, it requires careful consideration of task similarity, dataset distribution, and model compatibility to avoid negative outcomes.

\section{Types of Transfer Learning}\label{sec5}
Transfer learning can be categorized into several types based on the level of similarity between the source and target tasks, as well as the type of knowledge transferred. Here are some common types of transfer learning in forest fire detection:

\subsection{Inductive Transfer}
Inductive transfer learning is a type of transfer learning that involves transferring knowledge from a source task to a target task, where the target task is similar to the source task but not identical. This type of transfer learning is useful when the source and target tasks share some common characteristics but also have some differences.

In the context of forest fire detection, inductive transfer learning can be used to adapt a model trained on a dataset of forest fires in one region to detect forest fires in another region. For example, a model trained on a dataset of forest fires in California can be used to detect forest fires in India, as both regions have similar forest types and fire patterns. However, the model may need to be fine-tuned to account for differences in climate, vegetation, and other environmental factors between the two regions.

\subsection{Unsupervised Learning}

Unsupervised transfer learning is a type of transfer learning that involves transferring knowledge from a source task to a target task, where the target task is unsupervised, meaning that there is no labeled data available. This type of transfer learning is useful when there is a large amount of unlabeled data available but not enough labelled data to train a model.

In the context of forest fire detection, unsupervised learning can be used to cluster satellite images of forest fires to identify patterns and features. For example, a model can be trained on a large dataset of unlabeled satellite images of forest fires and then used to cluster them into different categories based on their characteristics. This can help to identify patterns and features that are indicative of forest fires and can be used to improve the performance of forest fire detection models.

\subsection{Transductive Transfer} Transductive transfer learning is a type of transfer learning that involves transferring knowledge from a source task to a target task, where the target task is similar to the source task, but the data distribution is different. This type of transfer learning is useful when the source and target tasks share some common characteristics, but the data distribution is different.

In the context of forest fire detection, transductive transfer learning can be used to adapt a model trained on a dataset of forest fires in one region to detect forest fires in another region with a different data distribution. For example, a model trained on a dataset of forest fires in a dry region can be used to detect forest fires in a humid region, as both regions have similar forest types and fire patterns, but the data distribution is different.

\subsection{Self-Taught Learning}
Self-taught learning is a type of transfer learning that involves transferring knowledge from a source task to a target task, where the target task is similar to the source task, but the data is unlabeled. This type of transfer learning is useful when there is a large amount of unlabeled data available but not enough labelled data to train a model.

In the context of forest fire detection, self-taught learning can be used to adapt a model trained on a dataset of labeled forest fires to detect forest fires in a new region with unlabeled data. For example, a model trained on a dataset of labeled forest fires in California can be used to detect forest fires in India, where there is a large amount of unlabeled data available.

\begin{table*}
\centering
\caption{Comparison of Different Types of Transfer Learning for Forest Fire Detection.}
\begin{tabular}{|p{1.5cm}|p{2.6cm}|p{2.6cm}|p{2.6cm}|p{2.6cm}|p{2.6cm}|} 
\hline
 & Inductive Transfer & Unsupervised Learning & Transductive Transfer & Self-Taught Learning & Multi-Task Learning \\ 
\hline
Source Task & Forest fires in one region & Labeled satellite images of forest fires & Forest fires in one region & Labeled forest fires & Forest fires and fire severity \\ 
\hline
Target Task & Forest fires in another region & Unlabeled satellite images of forest fires & Forest fires in another region with different data distribution & Forest fires in a new region with unlabeled data & Forest fires and fire severity \\ 
\hline
Data Distribution & Same & Same & Different & Unlabeled & Same \\ 
\hline
Knowledge Transferred & Features, patterns & Features, patterns & Features, patterns & Features, patterns & Features, patterns \\ 
\hline
Advantages & Adapts to new regions with minimal data & Adapts to new regions with no labeled data & Adapts to new regions with different data distributions & Adapts to new regions with unlabeled data & Improves performance on multiple tasks \\ 
\hline
\end{tabular}
\end{table*}

\subsection{Multi-Task Learning} Multi-task learning is a type of transfer learning that involves training a single model on multiple tasks simultaneously, where the tasks are related. This type of transfer learning is useful when multiple related tasks need to be performed, and the model can benefit from learning the relationships between the tasks.

In the context of forest fire detection, multi-task learning can be used to train a model to detect forest fires and predict fire severity simultaneously. For example, a model can be trained on a dataset of forest fire images to detect the presence of fires and predict the severity of the fires. This can be done by sharing the same feature extractor for both tasks and having separate output layers for each task.

\section{Utilizing Pre-trained Models for Efficient Transfer Learning} \label{sec6}

pre-trained models play a critical role in developing machine learning applications, especially in scenarios like forest fire detection, where obtaining labelled data can be challenging and expensive. These models are neural networks that have already been trained on large-scale datasets, enabling them to capture essential features such as shapes, patterns, and textures that can be transferred to new tasks. In the context of forest fire detection, pre-trained models can be particularly useful. For example, a model pre-trained on a large dataset of satellite images can be fine-tuned to detect forest fires in a specific region. This can be achieved by leveraging the knowledge the model has already learned about recognizing patterns and features in satellite images and adapting it to the specific task of forest fire detection. The benefits of using pre-trained models in forest fire detection include:

\begin{itemize}
    \item \textbf{Leverage from large datasets:} Pre-trained models are trained on vast amounts of data, often in domains like image recognition or language processing. In the case of forest fire detection, a pre-trained model could have been trained on satellite imagery from various regions around the world. This foundational knowledge allows the model to generalize well when applied to new datasets with similar characteristics, making it capable of recognizing diverse vegetation types, soil conditions, and climatic factors that may influence fire behaviour.

    \item \textbf{Faster training times:} Because these models have already learned to identify crucial features, they require significantly less time to fine-tune on specific tasks. For instance, adopting a pre-trained model for detecting forest fires can lead to faster deployment in real-world applications, as less computational power and time are needed compared to training a model from scratch. This accelerated training process is especially beneficial in urgent scenarios, such as during fire seasons, where timely detection can mitigate the risk of extensive damage.

    \item \textbf{Reduced need for labeled data:} pre-trained models allow researchers and practitioners to work effectively with smaller labeled datasets. For example, if there are limited labelled satellite images of forest fires in a particular region, a pre-trained model can be fine-tuned using this small dataset. This makes it possible to achieve high performance without the need for extensive labeled data, thereby lowering the barrier to entry for developing sophisticated fire detection systems in regions where data collection is difficult.

    \item \textbf{Improved performance on related tasks:} Utilizing pre-trained models often results in enhanced performance compared to training models from scratch. In the context of forest fire detection, models pre-trained on diverse datasets can recognize subtle patterns in satellite images that indicate the presence of a fire. These may include thermal anomalies, changes in vegetation health, or unusual smoke patterns. As a result, pre-trained models can significantly improve accuracy and response times for fire detection systems, enabling quicker interventions and better resource allocation during fire emergencies.

    \item \textbf{Adaptability to new conditions:} Pre-trained models can be fine-tuned to adapt to new environmental conditions, such as changes in vegetation due to seasonal shifts or climate variations. This adaptability is crucial in forest fire detection, as the model can learn to identify fire indicators in different landscapes and ecosystems.

    \item \textbf{Robustness to noise:} Models trained on large datasets often exhibit greater robustness to noise and variability in data. For forest fire detection, this means that pre-trained models can better handle variations in image quality or atmospheric conditions, leading to more reliable detection in challenging scenarios.

    \item \textbf{Cross-domain applicability:} Pre-trained models can often be adapted for different but related tasks. For example, a model trained for detecting forest fires can be adjusted for applications in monitoring agricultural burnings or urban fire incidents, maximizing the utility of existing resources.
\end{itemize}

Some examples of pre-trained models that can be used for forest fire detection include:
\begin{itemize}
    \item \textbf{Convolutional Neural Networks (CNNs):} These models are particularly well-suited for image recognition tasks and can be pre-trained on large datasets of satellite images. CNNs are capable of automatically extracting features from images, making them highly effective for detecting the visual signatures of forest fires.

    \item \textbf{Residual Networks (ResNets):} These models are known for their ability to learn complex patterns in data and can be pre-trained on extensive datasets of satellite images. ResNets utilize skip connections to mitigate issues like vanishing gradients, allowing them to train deeper networks that can capture more intricate details relevant to fire detection.

    \item \textbf{U-Net:} This model is a type of CNN that is particularly well-suited for image segmentation tasks and can be pre-trained on large datasets of satellite images. U-Net's architecture, which includes an encoder-decoder structure, enables precise localization of fire areas in satellite images, thus improving the accuracy of fire detection and assessment.

    \item \textbf{Inception Networks:} These networks leverage multiple filter sizes to capture various spatial features and are pre-trained on large image datasets such as ImageNet. They can be adapted for detecting fire by focusing on the characteristic colors and shapes associated with flames and smoke in satellite imagery.

    \item \textbf{VGGNet:} Known for its deep architecture, VGGNet is effective at image classification and can be pre-trained on large datasets. It can be fine-tuned for detecting forest fires by analyzing pixel-level details that indicate fire activity, making it useful in monitoring and risk assessment.

    \item \textbf{MobileNet:} A lightweight model ideal for real-time applications on mobile devices, MobileNet can be pre-trained on extensive datasets and adapted for forest fire detection in scenarios where computational resources are limited, such as in remote monitoring stations.
\end{itemize}

\section{Steps in the Transfer Learning Process}\label{sec7}
The following steps outline the transfer learning process, which allows researchers to effectively adapt pre-trained models for specific tasks such as forest fire detection, ensuring efficient use of existing knowledge while minimizing the need for extensive labeled data.
\subsection{Selecting a Suitable pre-trained Model} 
Selecting a suitable pre-trained model is a critical step in the transfer learning process for forest fire detection. The chosen model should align with the specific requirements of the task and be trained on datasets that include diverse features relevant to forest fire detection. Here are some key considerations to keep in mind when selecting a pre-trained model:

\begin{itemize}
    \item \textbf{Dataset Similarity:} The pre-trained model should be trained on a dataset that is similar to the dataset used for forest fire detection. For example, if the dataset for forest fire detection consists of satellite images, the pre-trained model should also be trained on satellite images.
    
 \item \textbf{Model Architecture:} The model architecture should be suitable for the task of forest fire detection. Different model architectures are better suited for different types of data and tasks. Here are some common model architectures and their suitability for forest fire detection:

\begin{enumerate}
    \item \textbf{Convolutional Neural Networks (CNNs):} CNNs are well-suited for image classification tasks, such as forest fire detection. They are designed to extract features from images and can learn to recognize patterns and objects within images.
    
    \item \textbf{Recurrent Neural Networks (RNNs):} RNNs are better suited for sequential data, such as time-series data or text data. They are not typically used for image classification tasks, but can be used for tasks such as predicting the spread of a forest fire over time.
    
    \item \textbf{Autoencoders:} Autoencoders are a type of neural network that can be used for unsupervised learning tasks, such as anomaly detection. They can be used to detect unusual patterns in images that may indicate a forest fire.
    
    \item \textbf{Generative Adversarial Networks (GANs):} GANs are a type of neural network that can be used for generating new images that are similar to a given dataset. They can be used to generate synthetic images of forest fires that can be used to augment a dataset.
\end{enumerate}
    
\item \textbf{Model Depth:} The depth of the model is also an important consideration. The depth of a model refers to the number of layers it has. Deeper models can learn more complex features, but may also be more prone to overfitting.

\begin{enumerate}
    \item \textbf{Shallow Models:} Shallow models have fewer layers and are less prone to overfitting. They are often used for simple tasks, such as binary classification. However, they may not be able to learn complex features and may not perform well on tasks that require a high level of accuracy.
    
    \item \textbf{Deep Models:} Deep models have more layers and can learn more complex features. They are often used for tasks that require a high level of accuracy, such as image classification. However, they may be more prone to overfitting and may require more data to train.
    
    \item \textbf{Very Deep Models:} Very deep models have many layers and can learn very complex features. They are often used for tasks that require a very high level of accuracy, such as object detection. However, they may be highly prone to overfitting and may require a large amount of data to train.
    
    \item \textbf{Residual Connections:} Residual connections are a technique used to train very deep models. They allow the model to learn residual functions, which can help to reduce the vanishing gradient problem and improve the model's ability to learn complex features.
    
    \item \textbf{Dense Connections:} Dense connections are a technique used to train very deep models. They allow the model to learn dense features, which can help to improve the model's ability to learn complex features.
\end{enumerate}

When selecting a model depth, it's essential to consider the following factors:

\begin{enumerate}
    \item \textbf{Task Complexity:} The complexity of the task will determine the required model depth. More complex tasks require deeper models.
    
    \item \textbf{Dataset Size:} The size of the dataset will determine the required model depth. Larger datasets can support deeper models.
    
    \item \textbf{Computational Resources:} The available computational resources will determine the required model depth. Deeper models require more computational resources.
    
    \item \textbf{Overfitting:} The risk of overfitting will determine the required model depth. Deeper models are more prone to overfitting.
\end{enumerate}

In the context of forest fire detection, a model depth of 10-20 layers is often sufficient. However, this can vary depending on the specific task and dataset. Here are some examples of models with different depths:

\begin{enumerate}
    \item \textbf{VGG16:} VGG16 is a deep model with 16 layers. It is well-suited for image classification tasks, such as forest fire detection.
    
    \item \textbf{ResNet50:} ResNet50 is a very deep model with 50 layers. It is well-suited for image classification tasks, such as forest fire detection.
    
    \item \textbf{U-Net:} U-Net is a deep model with 23 layers. It is well-suited for image segmentation tasks, such as distinguishing between fire-affected areas and surrounding vegetation.
\end{enumerate}
\item \textbf{Model Width:} The model's width, or the number of parameters, is also important. The width of a model refers to the number of neurons in each layer, which determines the number of parameters in the model. Wider models can learn more complex features but may also be more computationally expensive.

\begin{enumerate}
    \item \textbf{Narrow Models:} Narrow models have fewer neurons in each layer and fewer parameters. They are less computationally expensive and may be faster to train, but may not be able to learn complex features.
    
    \item \textbf{Wide Models:} Wide models have more neurons in each layer and more parameters. They can learn more complex features but may be more computationally expensive and may require more data to train.
    
    \item \textbf{Very Wide Models:} Very wide models have many neurons in each layer and many parameters. They can learn very complex features but may be highly computationally expensive and may require a large amount of data to train.
\end{enumerate}

When selecting a model width, it's essential to consider the following factors:

\begin{enumerate}
    \item \textbf{Task Complexity:} The complexity of the task will determine the required model width. More complex tasks require wider models.
    
    \item \textbf{Dataset Size:} The size of the dataset will determine the required model width. Larger datasets can support wider models.
    
    \item \textbf{Computational Resources:} The available computational resources will determine the required model width. Wider models require more computational resources.
    
    \item \textbf{Overfitting:} The risk of overfitting will determine the required model width. Wider models are more prone to overfitting.
\end{enumerate}

In the context of forest fire detection, a model width of 64-256 neurons per layer is often sufficient. However, this can vary depending on the specific task and dataset.

Some examples of models with different widths:

\begin{enumerate}
    \item \textbf{VGG16:} VGG16 is a model with a width of 64-512 neurons per layer. It is well-suited for image classification tasks, such as forest fire detection.
    
    \item \textbf{ResNet50:} ResNet50 is a model with a width of 64-1024 neurons per layer. It is well-suited for image classification tasks, such as forest fire detection.
    
    \item \textbf{U-Net:} U-Net is a model with a width of 64-256 neurons per layer. It is well-suited for image segmentation tasks, such as distinguishing between fire-affected areas and surrounding vegetation.
\end{enumerate}
    
  \item \textbf{pre-training Task:} The pre-training task should be similar to the task of forest fire detection. The pre-training task determines the type of features that the model learns, and a similar task can provide more relevant features for forest fire detection. For example:

\begin{enumerate}
    \item \textbf{Image Classification:} If the pre-trained model was trained on an image classification task, it may be well-suited for forest fire detection. Image classification tasks involve learning features that are relevant for distinguishing between different classes of images, which can be similar to distinguishing between fire-affected and unaffected areas.
    
    \item \textbf{Object Detection:} If the pre-trained model was trained on an object detection task, it may be well-suited for forest fire detection. Object detection tasks involve learning features that are relevant for detecting objects within images, which can be similar to detecting fire-affected areas within images.
    
    \item \textbf{Segmentation:} If the pre-trained model was trained on a segmentation task, it may be well-suited for forest fire detection. Segmentation tasks involve learning features that are relevant for distinguishing between different regions within images, which can be similar to distinguishing between fire-affected and unaffected areas.
\end{enumerate}

\item \textbf{pre-training Dataset Size:} The size of the pre-training dataset is also important. A larger pre-training dataset can provide more robust features but may also be more computationally expensive. For example:

\begin{enumerate}
    \item \textbf{Small Datasets:} Small datasets may not provide enough data to learn robust features but may be faster to train.
    
    \item \textbf{Medium Datasets:} Medium-sized datasets may provide a good balance between feature robustness and computational expense.
    
    \item \textbf{Large Datasets:} Large datasets may provide very robust features but may be computationally expensive to train.
\end{enumerate}

\item \textbf{Model Performance:} The performance of the pre-trained model on its original task is also an important consideration. A model that performs well on its original task is more likely to perform well on the task of forest fire detection. For example:

\begin{enumerate}
    \item \textbf{High-Performing Models:} Models that perform well on their original task may be more likely to perform well on the task of forest fire detection.
    
    \item \textbf{Low-Performing Models:} Models that perform poorly on their original task may not be suitable for forest fire detection.
\end{enumerate}

When selecting a pre-trained model, it's essential to consider the pre-training task, pre-training dataset size, and model performance. A model that is well-suited for forest fire detection should have a similar pre-training task, a large enough pre-training dataset, and good performance on its original task. Here are some examples of pre-trained models that may be suitable for forest fire detection:

\begin{enumerate}
    \item \textbf{VGG16:} VGG16 is a pre-trained model that was trained on an image classification task. It has a large pre-training dataset and good performance on its original task, making it a good candidate for forest fire detection.
    
    \item \textbf{ResNet50:} ResNet50 is a pre-trained model that was trained on an image classification task. It has a large pre-training dataset and good performance on its original task, making it a good candidate for forest fire detection.
    
    \item \textbf{U-Net:} U-Net is a pre-trained model that was trained on a segmentation task. It has a medium-sized pre-training dataset and good performance on its original task, making it a good candidate for forest fire detection.
\end{enumerate}

\end{itemize}
\begin{table*}[h]
\centering
\caption{Key Considerations for Selecting a Suitable pre-trained Model for Forest Fire Detection}
\begin{tabular}{|p{3cm}|p{4cm}|p{4cm}|p{4cm}|}
\hline
\textbf{Model Characteristic} & \textbf{Description} & \textbf{Considerations} & \textbf{Examples} \\
\hline
\textbf{Dataset Similarity} & The pre-trained model should be trained on a dataset similar to the dataset used for forest fire detection. & Task complexity, dataset size, computational resources & Satellite images, aerial images \\
\hline
\textbf{Model Architecture} & The model architecture should be suitable for the task of forest fire detection. & Task complexity, dataset size, computational resources & CNNs, RNNs, Autoencoders, GANs \\
\hline
\textbf{Model Depth} & The depth of the model is important for learning complex features. & Task complexity, dataset size, computational resources, overfitting & VGG16 (16 layers), ResNet50 (50 layers), U-Net (23 layers) \\
\hline
\textbf{Model Width} & The width of the model is important for learning complex features. & Task complexity, dataset size, computational resources, overfitting & VGG16 (64-512 neurons), ResNet50 (64-1024 neurons), U-Net (64-256 neurons) \\
\hline
\textbf{pre-training Task} & The pre-training task should be similar to the task of forest fire detection. & Task complexity, dataset size, computational resources & Image classification, object detection, segmentation \\
\hline
\textbf{pre-training Dataset Size} & The size of the pre-training dataset is important for learning robust features. & Task complexity, dataset size, computational resources & Small datasets, medium datasets, large datasets \\
\hline
\textbf{Model Performance} & The performance of the pre-trained model on its original task is important for forest fire detection. & Task complexity, dataset size, computational resources & High-performing models, low-performing models \\
\hline
\end{tabular}
\end{table*}

\subsection{Modifying the Model Architecture for Forest Fire Detection}
After selecting a pre-trained model, the next step is to modify its architecture to tailor it to the forest fire detection task using satellite images. This involves several critical modifications to optimize the model for effectively detecting forest fires from satellite imagery.

\subsubsection{Changing Input Dimensions to Accommodate Satellite Images}

The first modification is to adjust the input dimensions of the model to match the resolution of the satellite images used for detecting forest fires. The pre-trained model may have been trained on images with different resolutions or aspect ratios, which can lead to discrepancies in feature extraction if not aligned properly. 

For example, if the satellite images have a resolution of 256x256 pixels and the pre-trained model was trained on images of 224x224 pixels, resizing becomes necessary. This adjustment can be implemented using various methods:
\begin{itemize}
    \item \textbf{Resizing Layers:} A resizing layer can be added at the beginning of the model to scale input images to the required dimensions without significant loss of information.
    \item \textbf{Cropping:} If maintaining the aspect ratio is crucial, cropping techniques can be used to focus on specific regions of the satellite images that are most relevant for fire detection.
    \item \textbf{Padding:} In cases where the aspect ratios differ significantly, padding can be applied to standardize input dimensions while preserving the image's essential features.
\end{itemize}

These adjustments ensure that the model effectively processes satellite images and captures the relevant features for fire detection.

\subsubsection{Adjusting the Number of Output Classes for Forest Fire Detection}

The next modification involves adjusting the number of output classes to align with the specific requirements of the forest fire detection task. The model should output classifications that reflect the task's goals, such as detecting the presence or absence of a fire.

\begin{itemize}
    \item \textbf{Binary Classification:} If the task is simply to identify whether a fire is present or not, the model can be modified to produce a binary output. This can be achieved by changing the final layer to have a single neuron with a sigmoid activation function.
    \item \textbf{Multi-Class Classification:} In scenarios where multiple classes are necessary, such as identifying different types of fires or levels of severity, the output layer should include multiple neurons corresponding to each class. A softmax activation function can then be employed to provide probabilities for each category.
    \item \textbf{Multi-Label Classification:} In some cases, a satellite image may indicate multiple fire events or conditions. The architecture can be adapted to allow for multi-label classification, wherein each output class can independently indicate the presence of a fire, enhancing the model's versatility.
\end{itemize}

Adjusting the output classes in this manner tailors the model’s predictions to meet the specific operational requirements for forest fire monitoring and response.

\subsubsection{Adding Specialized Layers for Forest Fire Detection with Satellite Images}

Finally, the architecture may need specialized layers to focus on features unique to forest fire detection, such as temperature spikes, smoke patterns, and other environmental indicators captured in satellite images.

\begin{itemize}
    \item \textbf{Convolutional Layers with Small Kernel Sizes:} Adding convolutional layers with smaller kernel sizes can enhance the model's ability to detect fine-grained features, such as localized temperature increases or smoke plumes. These features are critical for accurately identifying fire presence in satellite imagery.
    \item \textbf{Attention Mechanisms:} Incorporating attention mechanisms can help the model focus on specific regions of interest, especially in complex images where fire features may be obscured by vegetation or other elements. Attention layers can weigh the importance of different parts of the image, enabling better detection of fire-related features.
    \item \textbf{Feature Pyramid Networks (FPN):} Implementing FPNs can improve the model's ability to detect objects at different scales, which is particularly useful in satellite images where the size of fire-affected areas can vary significantly.
    \item \textbf{Dropout Layers for Regularization:} To prevent overfitting, particularly when working with a smaller labeled dataset, dropout layers can be added after certain layers. This approach encourages the model to generalize better to unseen data, thus enhancing its robustness in practical applications.
\end{itemize}

By modifying the model architecture in these ways, the model can be effectively tailored to meet the specific demands of forest fire detection using satellite images. This tailored approach enables the model to learn and recognize features that are critical for the task, thereby improving its accuracy and overall effectiveness in detecting forest fires in real-time scenarios.

\subsection{Fine-Tuning the Model for Forest Fire Detection}

After modifying the model architecture to accommodate the specific requirements of forest fire detection with satellite images, the next step is to fine-tune the model on a new dataset specifically related to forest fires. This process is critical for enabling the model to learn from the new data and improving its performance in detecting forest fires.

\subsubsection{Key Considerations for Fine-Tuning the Model}

Fine-tuning the model involves training the adapted model on a new dataset consisting of satellite imagery that is labeled for instances of fire.

One key consideration is the learning rate. It is essential to use a lower learning rate during this process to ensure that the pre-trained weights are preserved while the model adapts to learn from the new data. A lower learning rate helps the model make subtle adjustments based on the new dataset without completely overwriting the valuable features it has already learned from the pre-trained model. A common strategy is to start with a smaller learning rate (e.g., $10^{-5}$) and adjust it based on the training dynamics.

Another important aspect is data augmentation. Implementing data augmentation techniques is vital to enhance the training dataset and make the model more robust against variations in image conditions. This is particularly crucial for satellite imagery, which can be influenced by various environmental factors such as lighting, weather conditions, and sensor noise. Common augmentation techniques include rotations and flips to simulate different perspectives, scaling and cropping to focus on various regions of interest, color adjustments to account for variations in satellite image acquisition, and adding Gaussian noise to enhance resilience against sensor imperfections.

The size and quality of the dataset used for fine-tuning the model are also critical factors that can significantly impact performance. A larger and more diverse dataset allows the model to learn a wider range of features, thereby improving its generalization ability. In the context of forest fire detection, it is important to ensure that the dataset contains various instances of fire in different conditions and from multiple regions to enhance the model's robustness.

Furthermore, attention should be paid to training time and computational resources. Fine-tuning the model can be computationally intensive and time-consuming, particularly when working with high-resolution satellite imagery and large datasets. It is essential to have sufficient computational resources, such as GPUs or TPUs, to expedite the training process. Additionally, monitoring the training process is crucial to avoid issues like overfitting or underfitting. Techniques such as early stopping, where training is halted when performance on a validation set begins to degrade, can be employed to optimize training duration.

Regular validation during the fine-tuning process is necessary to evaluate the model’s performance on unseen data. It’s important to set aside a validation dataset that is representative of the task. Metrics such as accuracy, precision, recall, and F1 score should be calculated to assess how well the model is performing in detecting forest fires. Additionally, using confusion matrices can provide insights into the types of errors the model is making.

Lastly, exploring different transfer learning strategies can enhance performance. For instance, progressively unfreezing layers of the pre-trained model during training can allow the model to adapt more effectively to the new dataset. Initially, only the last few layers can be trained, and as training progresses, earlier layers can be gradually unfrozen to allow for more feature-specific learning.

By carefully considering these factors and systematically fine-tuning the model on a relevant dataset, it is possible to adapt the model to learn effectively from the new data. This approach significantly enhances its performance in the crucial task of forest fire detection using satellite images. The fine-tuning process allows the model to leverage the foundational knowledge from the pre-trained model while also learning to detect unique features associated with forest fires, ultimately improving accuracy and response times in real-world applications.

\begin{figure*}
    \centering
    \includegraphics[scale=0.36]{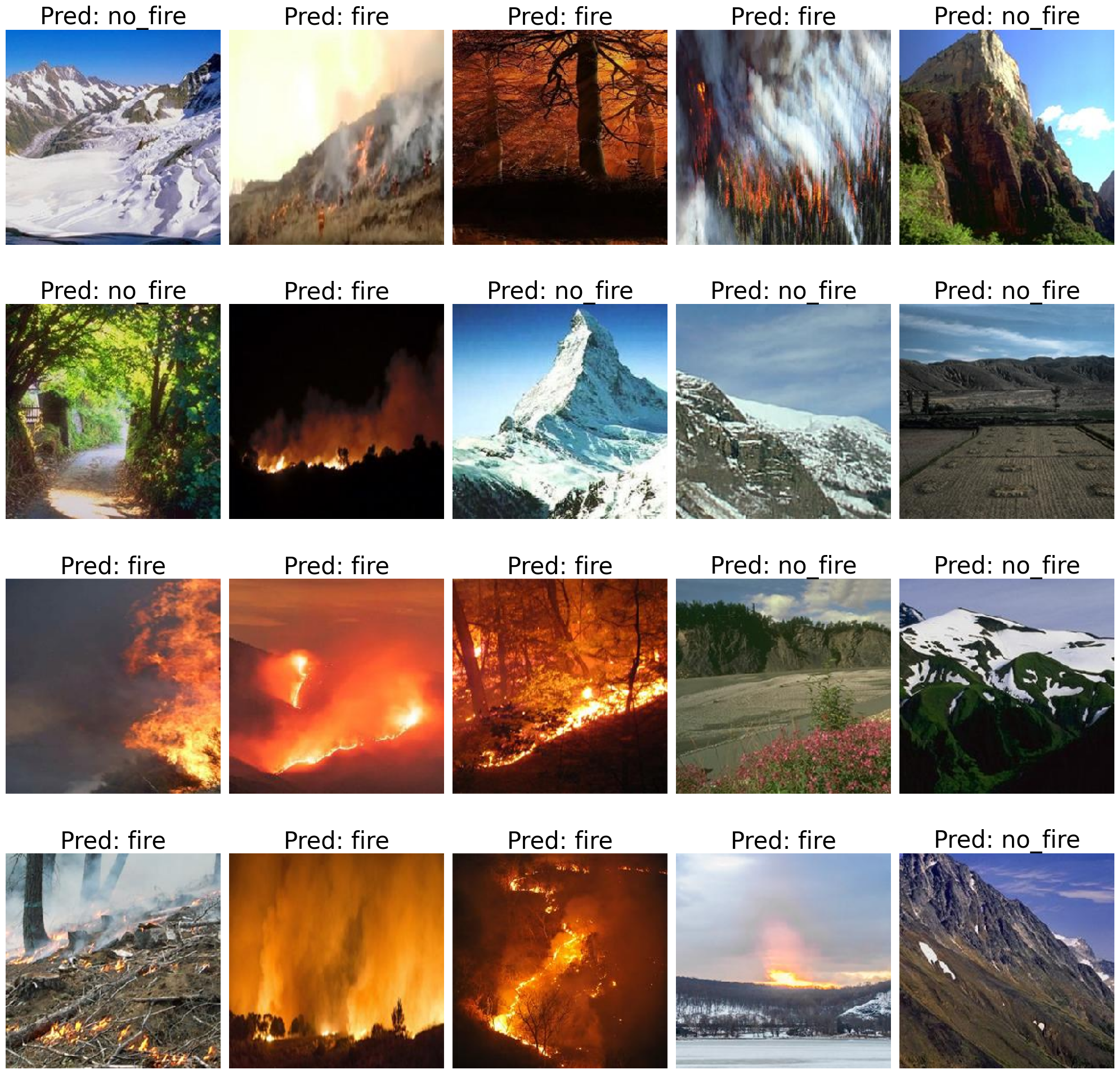}
    \caption{A visual illustration of the proposed model on predicting the condition of fire.}
    \label{result1}
\end{figure*}

\section{Experimental Results}\label{sec8}
In this section, we present the experimental results obtained from training and evaluating a deep learning model for forest fire detection using a pre-trained MobileNetV2 architecture.

\subsection{Dataset Description}
The dataset for our experiments comprises images balanced between two classes: \textit{fire} and \textit{no-fire}~\cite{ref1}. Each image has a resolution of in 3-channel RGB format. Images were gathered through extensive searches using various terms across multiple search engines. To ensure quality, we reviewed and cropped inappropriate elements, such as people and fire-extinguishing equipment, focusing solely on relevant fire regions. Additionally, we included images from the Uttarakhand forest fire dataset to enhance diversity and robustness. The dataset was divided into 80\% for training and 20\% for testing to effectively evaluate the model's performance.

\subsection{Model Architecture}
The model utilized is based on the MobileNetV2 architecture, which was fine-tuned for binary classification of images into two categories: 'fire' and 'no fire'. The architecture was adapted by removing the top layer of MobileNetV2 and adding a custom classification head comprising a global average pooling layer, a fully connected layer with 128 neurons and ReLU activation, followed by a dropout layer to mitigate overfitting. The final layer employs a sigmoid activation function for binary classification.

\subsection{Data Preparation}

The dataset was organized into three directories: \texttt{train}, \texttt{validation}, and \texttt{test}, each containing subfolders for each class. The data was augmented using techniques such as rotation, zooming, and horizontal flipping to improve model robustness. The images were resized to \(224 \times 224\) pixels and normalized.

\subsection{Training Process}

The model was compiled using the Adam optimizer with a binary cross-entropy loss function. The training was conducted over 200 epochs with a batch size of 32. The training and validation accuracy, as well as the loss, were recorded throughout the training process.

\subsection{Results}
Fig.~\ref{result1} shows a grid of predictions made by the MobileNetV2-based model for forest fire detection. Each image is labeled with the predicted class, either "fire" or "no fire," based on the model's inference.

The top row contains images of various landscapes, including snowy mountains and forests, with the model correctly identifying most of the "no fire" instances, except for one image in the middle, which shows a fire. Similarly, other rows display a mix of fire and non-fire scenarios. In some cases, such as in the second and third rows, the model correctly distinguishes between fire and non-fire in scenes with varying lighting conditions and settings (e.g., dense forests).

In the final row, the model continues to perform well, accurately detecting most fire instances. However, there are occasional misclassifications, such as identifying fire in an image with a sunset in the background (right column), which may be due to visual similarities between the color of fire and sunset. Overall, the model demonstrates strong performance in differentiating between fire and non-fire scenarios, though occasional misclassifications highlight some challenges in distinguishing similar visual features (e.g., fire vs. bright sources).

\begin{figure}[h]
    \centering
    \includegraphics[width=0.3\textwidth]{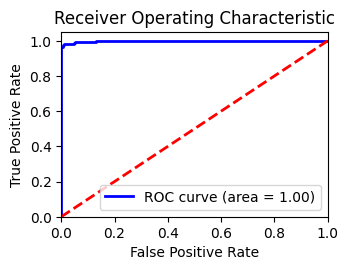}
    \caption{Illustration of Receiver Operating Curve (ROC).}
    \label{fig:roc}
\end{figure}

\begin{figure}[h]
    \centering
    \includegraphics[width=0.3\textwidth]{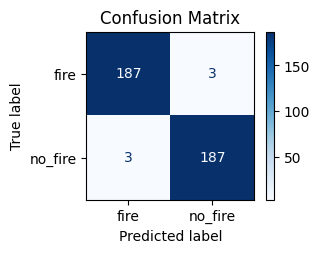}
    \caption{Illustration of the confusion matrix.}
    \label{fig:conf}
\end{figure}

Figures~\ref{fig:roc} and \ref{fig:conf} illustrate the ROC and confusion matrix, respectively. The Receiver Operating Characteristic (ROC) curve provides a visual representation of the model's classification performance across various threshold settings. It plots the true positive rate (TPR or sensitivity) against the false positive rate (FPR) at different decision thresholds. A perfect classifier would have a ROC curve that passes through the top-left corner, indicating high sensitivity and low false-positive rate. The Area Under the Curve (AUC) is commonly used to summarize the ROC, where a value close to 1 indicates excellent model performance.

The Confusion Matrix offers a detailed breakdown of the model's predictions, highlighting the counts of true positives, true negatives, false positives, and false negatives. This matrix allows for deeper insight into the types of errors the model makes. For instance, false positives (predicting "fire" when there is none) or false negatives (failing to detect a fire) are critical in the context of forest fire detection, where minimizing both error types is essential for safety and reliability. By analyzing the confusion matrix, we can identify any biases in the model and fine-tune them to improve accuracy and reduce misclassification rates.

\begin{figure}[h]
    \centering
    \includegraphics[width=0.4\textwidth]{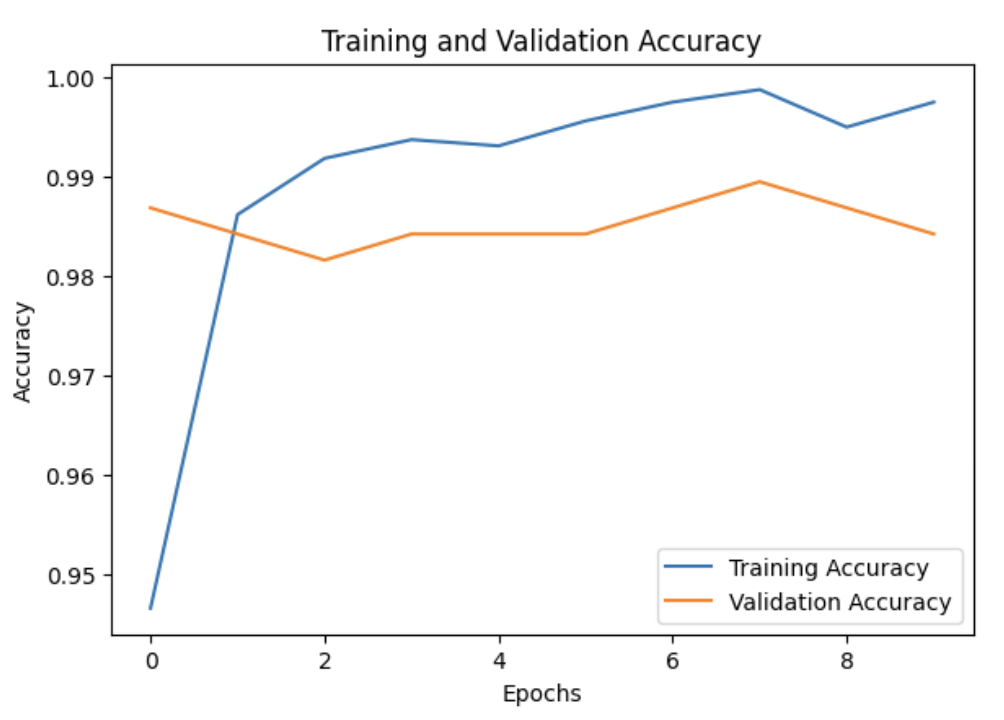}
    \caption{Training and validation accuracy and loss over epochs.}
    \label{fig:accuracy}
\end{figure}

\begin{figure}[h]
    \centering
    \includegraphics[width=0.4\textwidth]{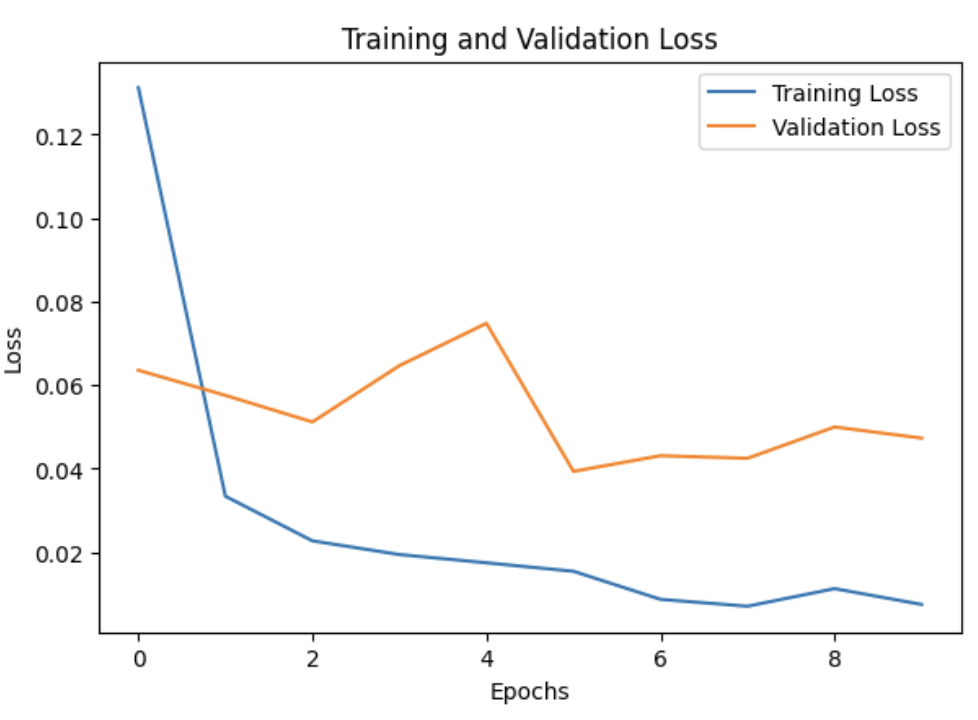}
    \caption{Training and validation accuracy and loss over epochs.}
    \label{fig:loss}
\end{figure}

Figs \ref{fig:accuracy} and \ref{fig:loss} illustrate the training and validation accuracy and loss, respectively.
The result illustrates the high performance of the MobileNetV2-based model for forest fire detection, with both training and validation accuracy exceeding 99\%. The training accuracy rapidly converges to nearly 100\% within the first few epochs, while validation accuracy improves more gradually, stabilizing around 99\% after the fourth epoch. This close alignment between training and validation accuracy suggests that the model generalizes well to unseen data, indicating minimal overfitting. The use of data augmentation and dropout during training likely contributed to this robust performance across diverse fire imagery.

The graph of training and validation loss shows that the model quickly reduces its training loss (blue curve), dropping sharply in the first two epochs and stabilizing near zero by epoch 6. This indicates that the model is highly effective in fitting the training data, with minimal errors as the training progresses. 

However, the validation loss (orange curve) fluctuates more, peaking around epoch 5 before slightly decreasing, though it remains consistently higher than the training loss. This suggests some overfitting, where the model performs better on the training data than on unseen validation data. Despite this, the overall downward trend toward the later epochs indicates the model is still generalizing well, though further regularization could potentially smooth out the validation loss fluctuations.

\section{Conclusion}\label{sec9}
In conclusion, our study demonstrates the efficacy of transfer learning in improving forest fire detection using deep learning models, specifically by fine-tuning pre-trained MobileNetV2 architecture. By leveraging the advantages of transfer learning, we were able to achieve high accuracy in detecting fires with limited labeled data, as demonstrated by the experimental results. The ROC curve and confusion matrix highlighted the model's strong performance in differentiating between fire and no-fire instances, with high true positive rates and minimal false positives. The dataset, enriched with images from the Uttarakhand forest fire, allowed us to adapt our model to real-world challenges such as diverse terrain and varying fire conditions. The graphical results further illustrate the model's robustness in classifying images accurately, even when confronted with complex and varied environments. These findings suggest that transfer learning offers a promising, scalable approach to forest fire detection in regions like India, where data scarcity and regional diversity pose unique challenges to traditional machine learning methods. Through this work, we have shown that deep learning, coupled with pre-trained models, can be effectively deployed to mitigate the devastating impacts of forest fires by facilitating timely and accurate detection.

\bibliographystyle{IEEEtran}
\bibliography{paper.bib}

\end{document}